\documentclass[letterpaper, 10 pt, conference]{ieeeconf}  

\IEEEoverridecommandlockouts                              

\usepackage{graphics} 
\usepackage[demo]{graphicx}
\usepackage{epsfig} 
\usepackage{times} 
\usepackage{amsmath} 
\usepackage{amssymb}  
\usepackage{hyperref}
\usepackage{algorithm}
\usepackage{algorithmic}
\usepackage{textcomp}
\usepackage{float}
\usepackage{algorithm}
\usepackage{algorithmic}
\usepackage{mathtools}
\usepackage{commath}
\usepackage{fixmath}
\usepackage{subcaption}
\usepackage{import}
\usepackage{bbm}
\usepackage{todonotes}
\usepackage[strings]{underscore}
\usepackage{setspace}
\let\Algorithm\algorithm
\renewcommand\algorithm[1][]{\Algorithm[#1]\setstretch{1.15}}
\usepackage{textcomp}

\title{\LARGE \bf
Recent Advances in Human-Robot Collaboration \\ Towards Joint Action
}

\author{
Yeshasvi Tirupachuri$^{1,2}$, Gabriele Nava$^{1,2}$, Lorenzo Rapetti$^{1,3}$, \\ Claudia Latella$^{1}$, Kourosh Darvish$^{1}$, Daniele Pucci$^{1}$
\thanks{This work is supported by \href{https://andy-project.eu/}{An.Dy} project which has received funding from the European Union\textquotesingle s Horizon 2020 Research and Innovation Programme under grant agreement No. 731540.}
\thanks{$^{1}$Dynamic Interaction Control, Italian Institute of Technology, Genoa, Italy {\tt\small name.surname@iit.it}}%
\thanks{$^{2}$DIBRIS, University of Genoa, Italy}
\thanks{$^{3}$Machine Learning and Optimisation, The University of Manchester,
 Manchester, United Kingdom.}
}

\begin{document}

\maketitle
\thispagestyle{empty}
\pagestyle{empty}

\begin{abstract}

Robots existed as separate entities till now, but the horizons of a symbiotic human-robot partnership are impending. Despite all the recent technical advances in terms of hardware, robots are still not endowed with desirable relational skills that ensure a social component in their existence. This article draws from our experience as roboticists in Human-Robot Collaboration (HRC) with humanoid robots and presents some of the recent advances made towards realizing intuitive robot behaviors and partner-aware control involving physical interactions.

\end{abstract}

\section{Introduction}
\label{introduction}


The current decade (2010-2020) will be etched in history as the decade of many key successes in the field of robotics in general. The success stories are marked by some of the most practical and interesting research work witnessed through government backed competitions like DARPA Robotics Challenge (DRC)~\cite{Krotkov:2017:DRC:3074644.3074647}\cite{doi:10.1177/0278364916688254}. Also, the total amount of venture capital investments toward robotic startups has been steadily increasing through out the decade which resulted in myriad of sophisticated products like Jibo, Mayfield Robotics Kuri, Anki Cozmo, Rethink Robotics Baxter, Franka Emika Panda, Universal Robotics U5, DJI Phantom. However, as the decade draws in, streams of failures dawn on may robotic startups leading to an eventual bankruptcy.

On careful analysis, one can conclude that most of the startups that failed attempted to provide social robots or promised higher emphasis on social component in their robots. While on the other hand, successful companies focused more on delivering products that are function oriented e.g., Franka Emika and Universal Robotics products are for collaborative manufacturing, DJI drones are for creative fields and entertainment.

An interesting thought exercise is to compare and contrast the smart devices market and robotics market. At the hardware level, the key enabling technologies behind smart devices and the current robots are similar e.g., interactive touch screens, high capacity graphics processing units, high bandwidth and low latency communication devices etc. However, the momentum sustained by smart devices and the rapid cultural adaptation of them in the previous decade is markedly more prominent than that of robots in the current decade. Although built as general purpose computing units, the promise and strength of smart devices lie in providing connectivity in the digital space, access to information and a myriad of tools to enhance creativity and productivity. In contrast, robots are expected to share the same physical space alongside humans and are expected to have many anthropomorphic traits ranging from physical structure to emotional intelligence that can assist and augment human life. So, in order to have more cultural acceptance and adoption, robots need to have relational skills that ensure a social component in their existence.

Businesses are capitalistic entities by nature and they always try to lower their costs and increase their profit margins. The global trend of increased wealth gap in the western society is a direct result of decades operating costs optimization by relocating the business operations from a labor expensive regions to inexpensive regions. Robotics and automation promises another level of reduction in operating costs by replacing as much human labor as possible and use robots instead, as they enable increased productivity. However, the key industries that are primed to be benefited through the recent technological advances in robotics are primarily logistics and manufacturing, where the labor replacement can be straight forward. Warehouse management through an autonomous fleet of robots and light weight collaborative robots for small and medium scale manufacturing enterprises are some of the emerging examples.

A fully autonomous humanoid robot that embodies all the human traits, from navigational capabilities to emotional capabilities, can be ascribed as the holy grail among roboticists. Service industries like hospitality and entertainment will have a big transformation as humanoid robot technology advances. Similarly, healthcare industry, in particular, elderly care and assistance will see a big improvement in catering the needs of rising elderly population across the world. Both verbal and non-verbal communication capabilities play a vital role to successfully employ humanoids in these sectors.

This article draws from our experience in Human-Robot Collaboration (HRC) with humanoid robots and presents some of the recent advances made towards realizing intuitive robot behaviours and partner-aware control involving physical interactions.

\section{Background}
\label{background}

\begin{figure}[t]
	\centering
	\includegraphics[scale=0.3]{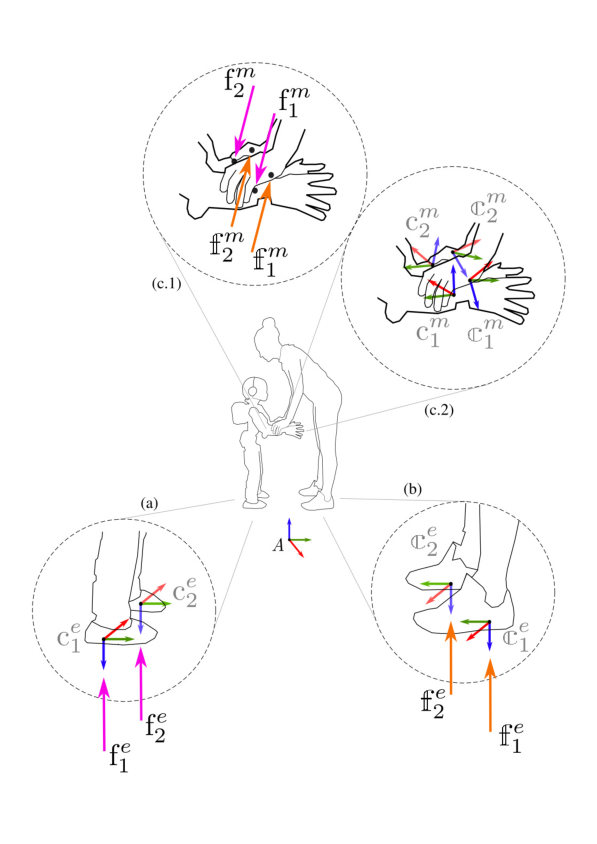}
	\caption{A typical human-robot dynamic interaction scenario}
	\label{pHRI}
\end{figure}

A typical Human-Robot Collaboration scenario is shown in Fig. \ref{pHRI}. There are two agents: the human, and the robot. Both agents are physically interacting with the environment and, in addition, are also engaged in interaction with each other. The contact locations and the interaction wrench exchanged at each contact location are highlighted.

Although the importance of nonverbal communication has been largely overestimated before \cite{mehrabian2017nonverbal}, it still plays a significant role in human interactions \cite{knapp2013nonverbal}\cite{beattie2004visible}. Accordingly, a robot's nonverbal behavior is very critical in increasing the trust in the human interacting with it and establish a more engaging communication between them towards a common goal. This can be manifested through some intuitive robot behaviors in response to the interaction with the human.
\section{Recent Developments}
\label{recent-developments}

Human habitats are dynamic environments in nature. Collisions pose a significant challenge to robots in these dynamic environments and the capacity to detect, isolate, identify and react is fundamental for their coexistence alongside humans \cite{haddadin2017robot}. Stability is one of the core capacities that is desirable in the case of humanoid robots and their ability to be robust to external perturbations is an active topic of research \cite{nava2016stability}. 

Physical interactions from a human during HRC are often intentional and can provide informative insights that can augment the task completion \cite{bajcsy2017learning}. Consider an example case of a robot moving its center of mass (CoM) along a given Cartesian reference trajectory to perform a complicated task of sit-to-stand transition i.e., stand-up task. An intuitive interaction of a human with the intention to speed up the robot motion is to apply forces in the robot's desired direction. Under such circumstances, traditionally, the robot can either render a compliant behavior through impedance or admittance control or be robust to any external interactions even if it is helpful for the task at hand. Instead, a more intuitive behavior is to advance further along the reference trajectory and stand-up quicker.

\begin{figure}[t]
    \centering
    \begin{subfigure}{0.125\textwidth}
        \centering
        \includegraphics[scale=0.10]{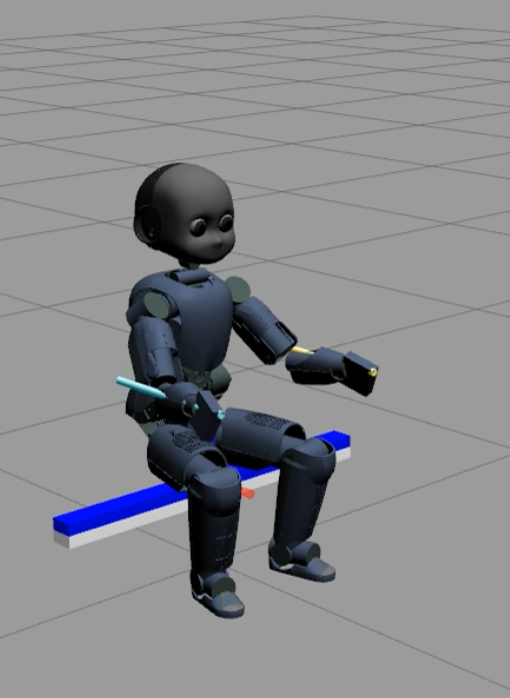}
        \caption{}
        \label{fig:state-1.png}
    \end{subfigure}%
    \begin{subfigure}{0.125\textwidth}
        \centering
        \includegraphics[scale=0.10]{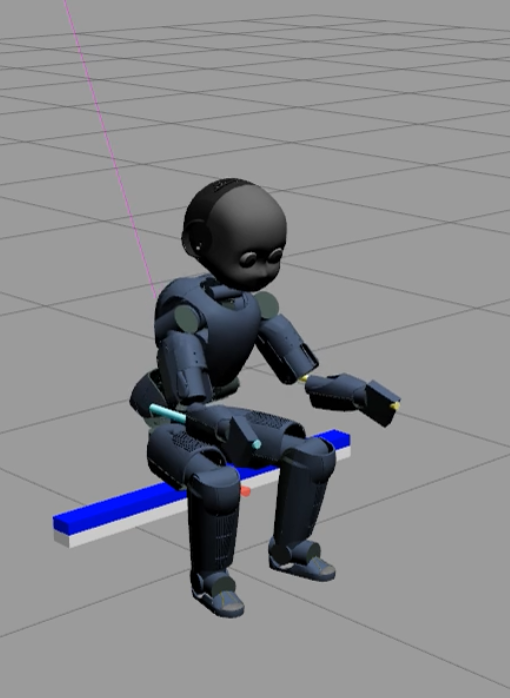}
        \caption{}
        \label{fig:state-2.png}
    \end{subfigure}%
    \begin{subfigure}{0.125\textwidth}
        \centering
        \includegraphics[ scale=0.10]{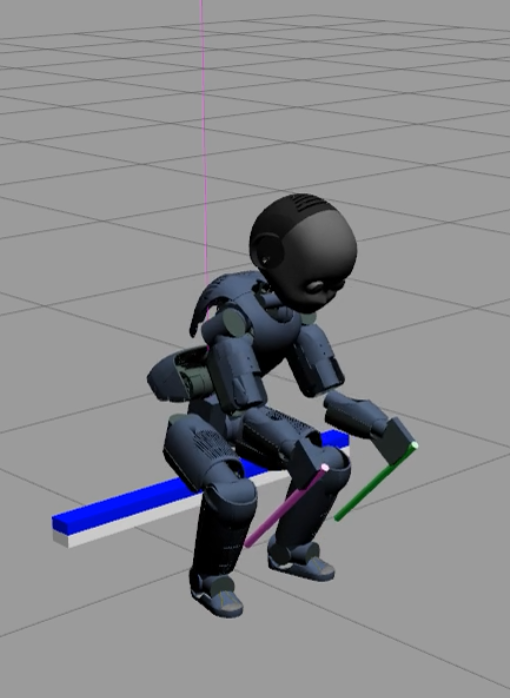}
        \caption{}
        \label{fig:state-3.png}
    \end{subfigure}%
    \begin{subfigure}{0.125\textwidth}
        \centering
        \includegraphics[scale=0.10]{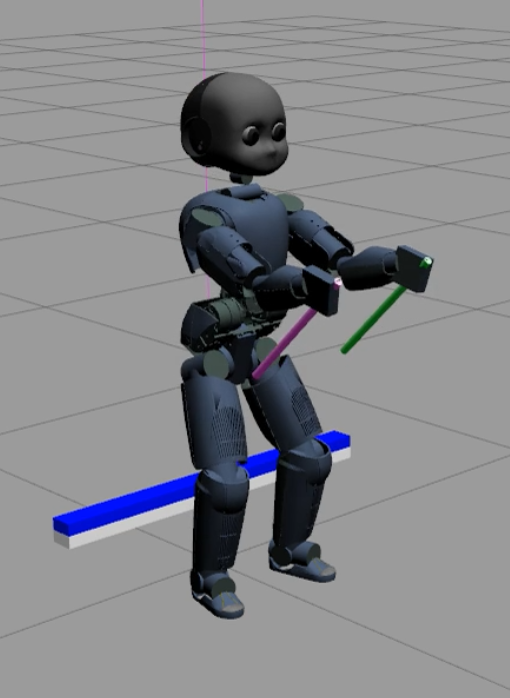}
        \caption{}
        \label{fig:state-4.png}
    \end{subfigure}
    \caption{iCub at different states during sit-to-stand transition}
    \label{fig:stand-up-states}
\end{figure}

\begin{figure}[t]
    \centering
    \includegraphics[scale=0.1]{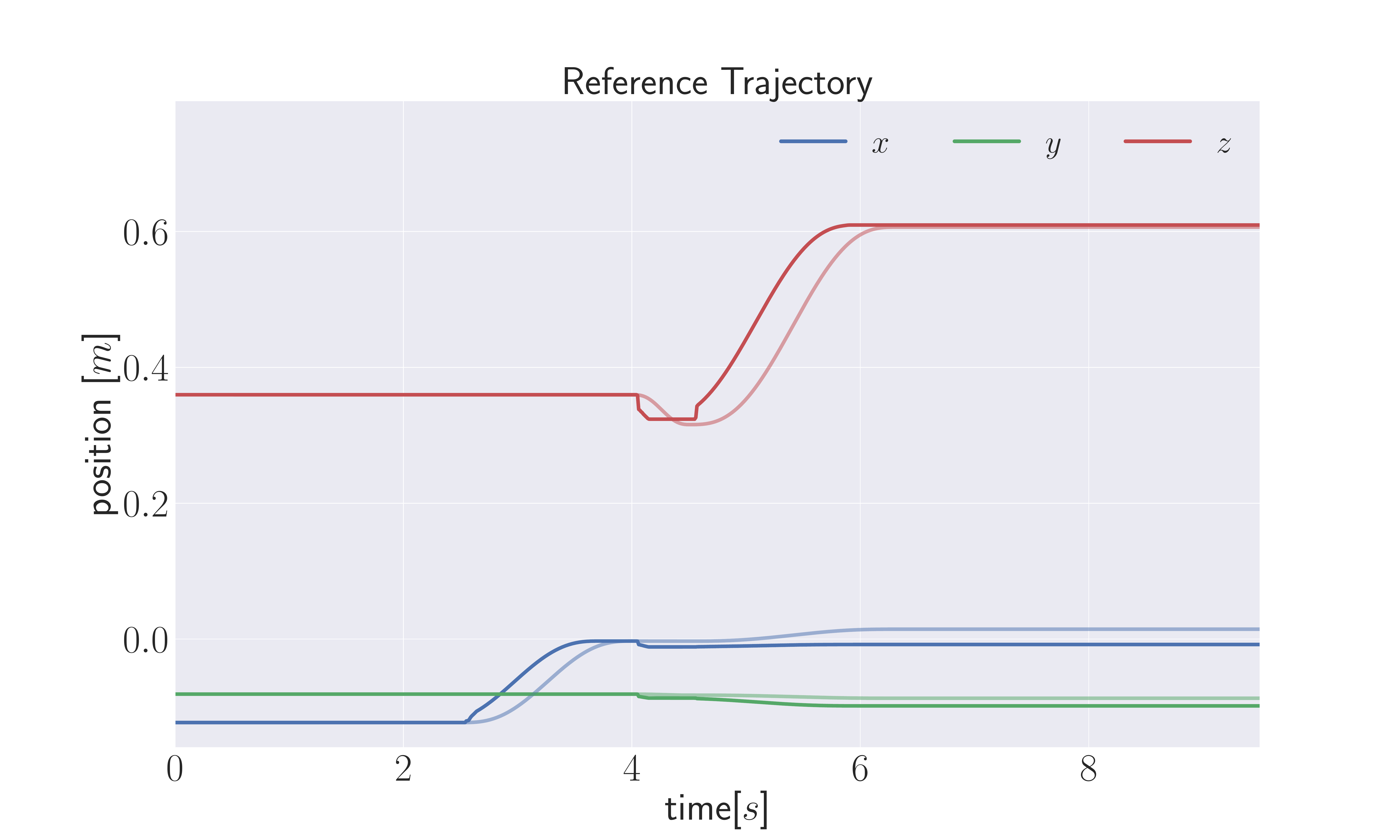}
    \caption{Center of Mass reference trajectory advancement}
    \label{fig:simulation-reference-trajectory}
\end{figure}

We recently proposed a trajectory advancement approach through which the robot can advance along the reference trajectory leveraging assistance from physical interactions \cite{trajectory-advancement}. Different stages of the robot stand-up task are highlighted in \ref{fig:stand-up-states} along with the interaction wrench at the hand that mimics human interaction for assistance. The CoM reference trajectory advancement using human assistance is shown in Fig.\ref{fig:simulation-reference-trajectory} along with the reference trajectory without trajectory advancement highlighted in reduced transparency.

The idea of body-schema has been greatly investigated by psychologists and neuroscientists \cite{berlucchi1997body}\cite{morasso2015revisiting}. Body-schema is a sensorimotor representation of the body that is fundamentally plastic in nature and it is crucial for the spatial organization of action through planning and guides the execution of movements. It also serves as a tool to explain human mastery of tool usage \cite{martel2016tool}. More importantly, body-schema not only serves as an individual's body representation but it is also important in representing bodies of others, facilitating nonverbal interpersonal communication \cite{chaminade2005fmri}. A few application-oriented research investigations of body-schema are conducted using robots \cite{Hoffmann:2010:BSR:2208749.2208799}\cite{vicente2016online}. 

Collaborative scenarios with a humanoid robotic agent are complex in nature. The dynamics of all the agents involved play a crucial role in shaping the interaction. So, simply considering the mathematical model of the robot to formulate control laws often fall short in understanding the interaction more concretely. Instead, the dynamics of all the agents involved have to be considered together rather than in isolation. Our recent work takes into account the dynamics of the combined system and present a coupled-dynamics formalism for collaboration scenarios \cite{tirupachuri2019towards}. Furthermore, we propose partner-aware robot control techniques that consider the joint torque quantities of the interacting robot towards accomplishing a shared goal. 

\begin{figure*}[t]
	\centering
	\begin{subfigure}{0.24\textwidth}
		\centering
		\includegraphics[scale=0.045]{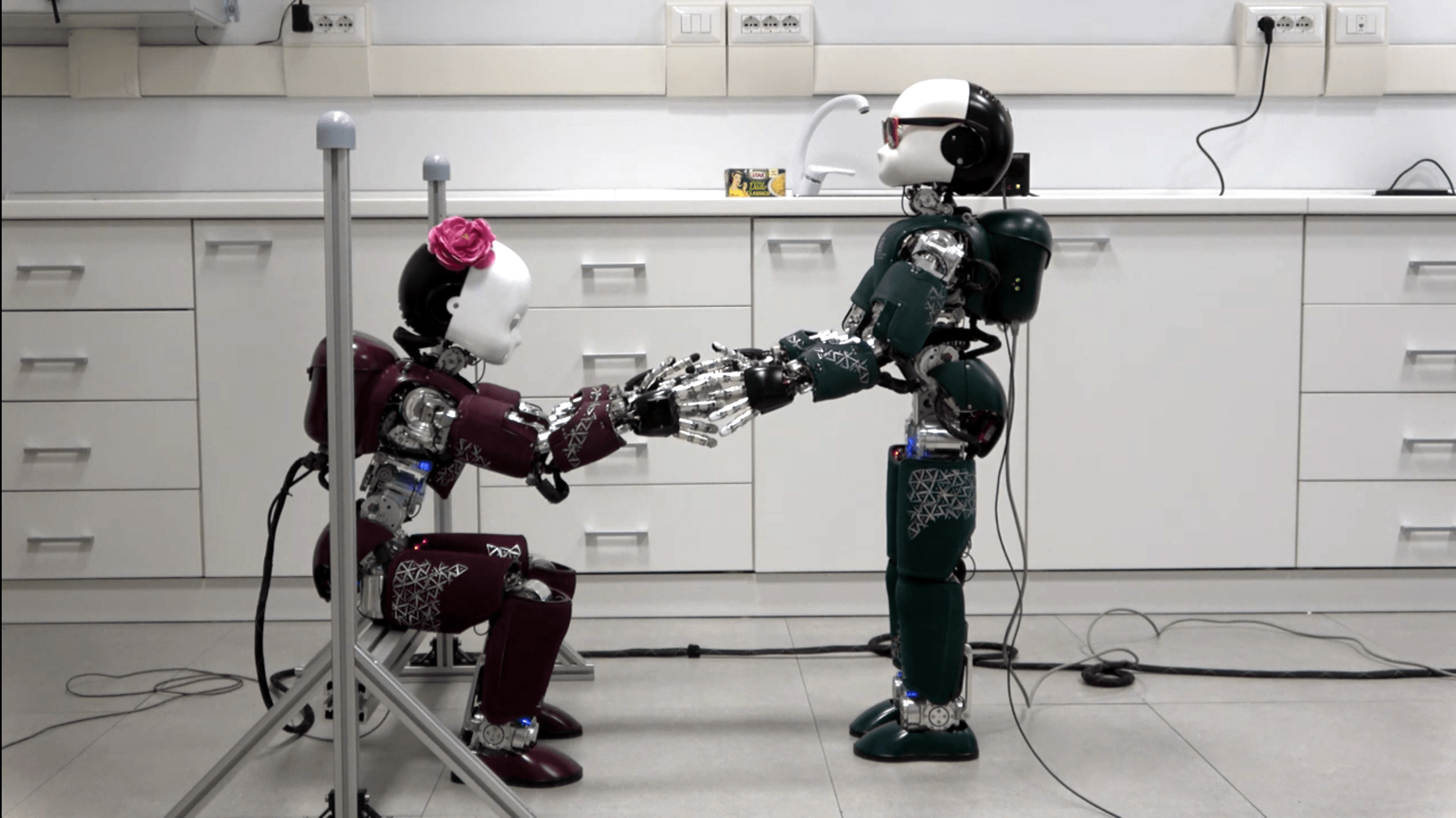}
		\caption{}
		\label{fig:state1}
	\end{subfigure}%
	\begin{subfigure}{0.24\textwidth}
		\centering
		\includegraphics[scale=0.045]{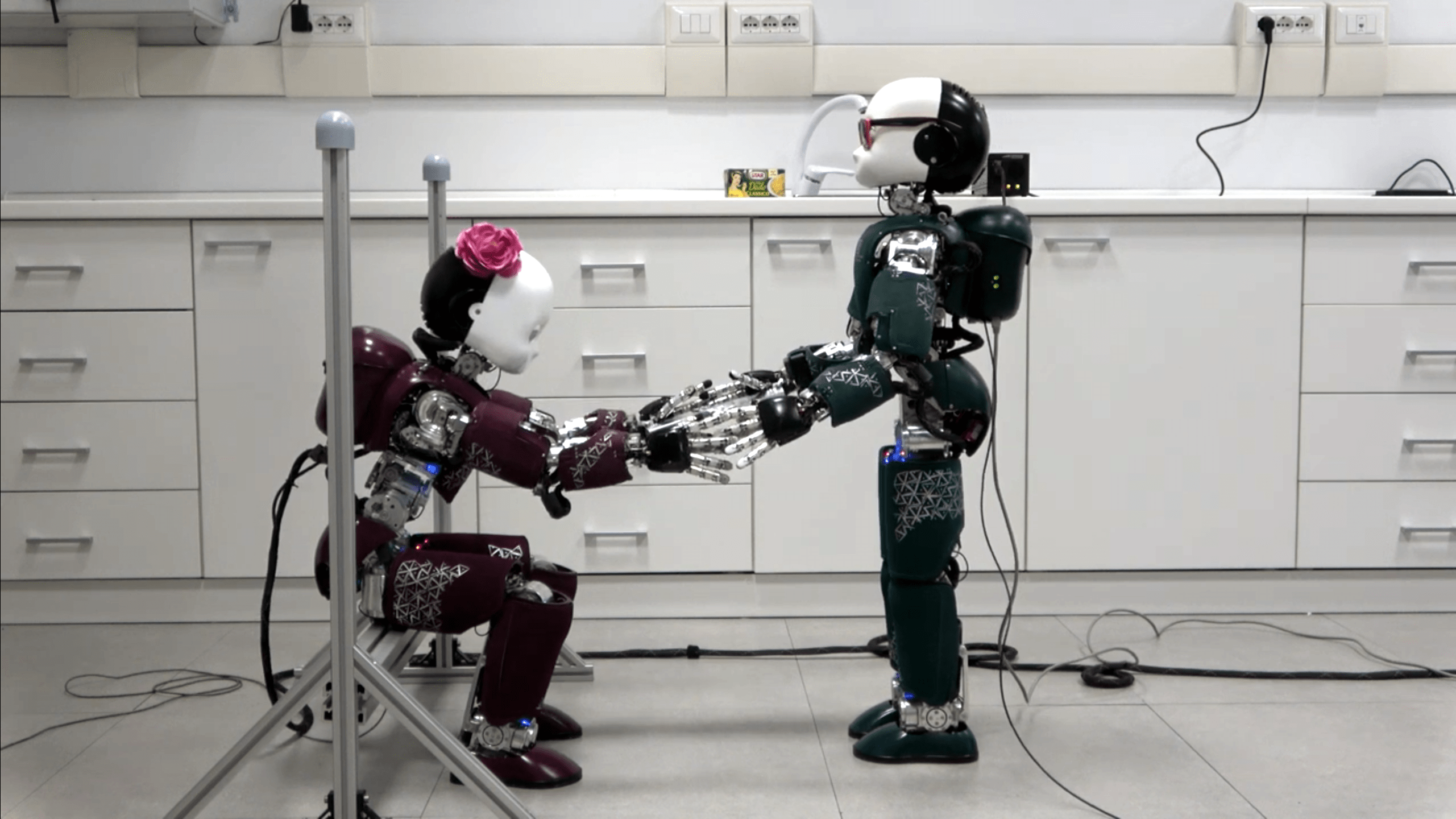}
		\caption{}
		\label{fig:state2}
	\end{subfigure}
	\begin{subfigure}{0.24\textwidth}
		\centering
		\includegraphics[scale=0.045]{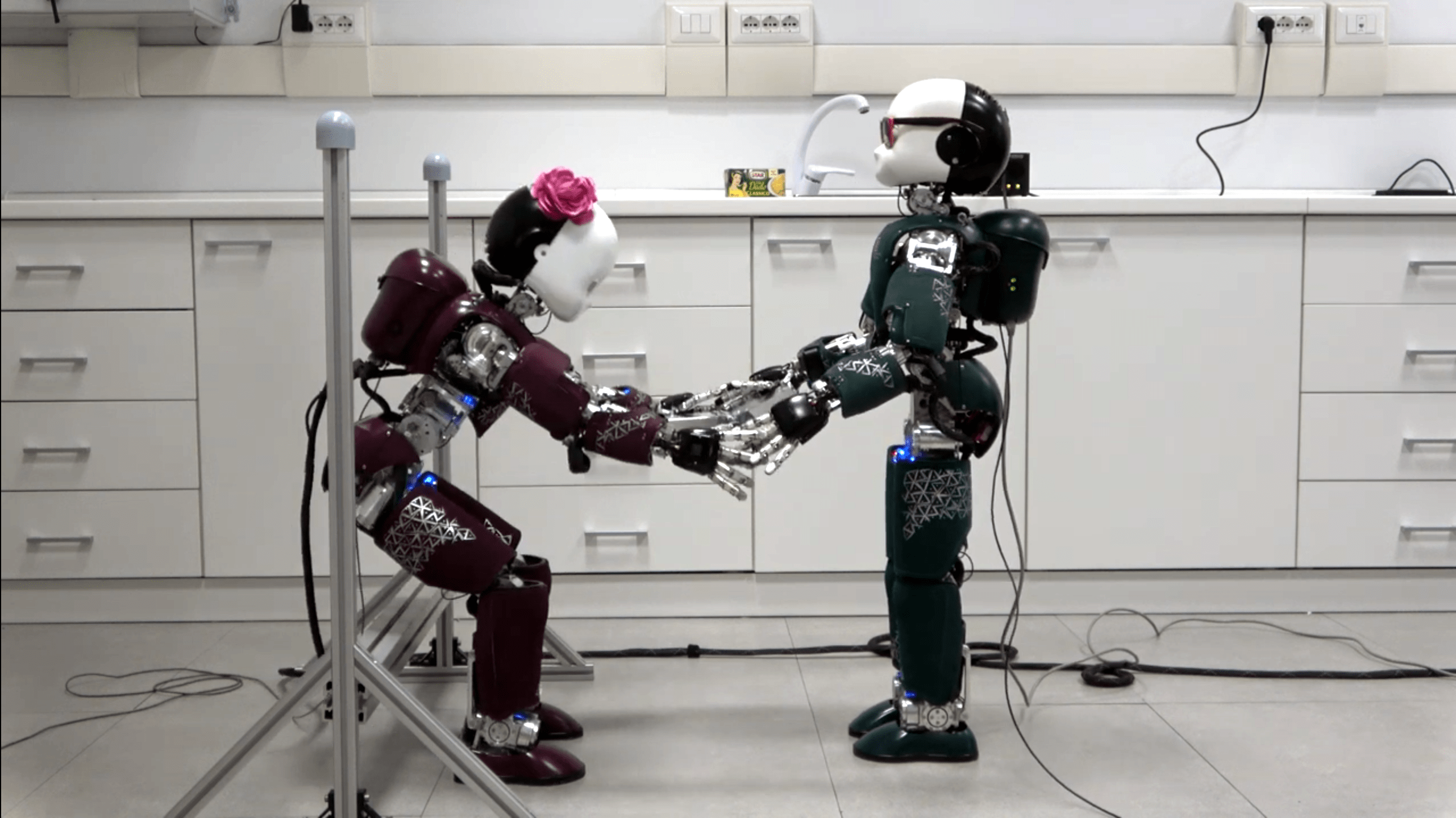}
		\caption{}
		\label{fig:state3}
	\end{subfigure}%
	\begin{subfigure}{0.24\textwidth}
		\centering
		\includegraphics[scale=0.045]{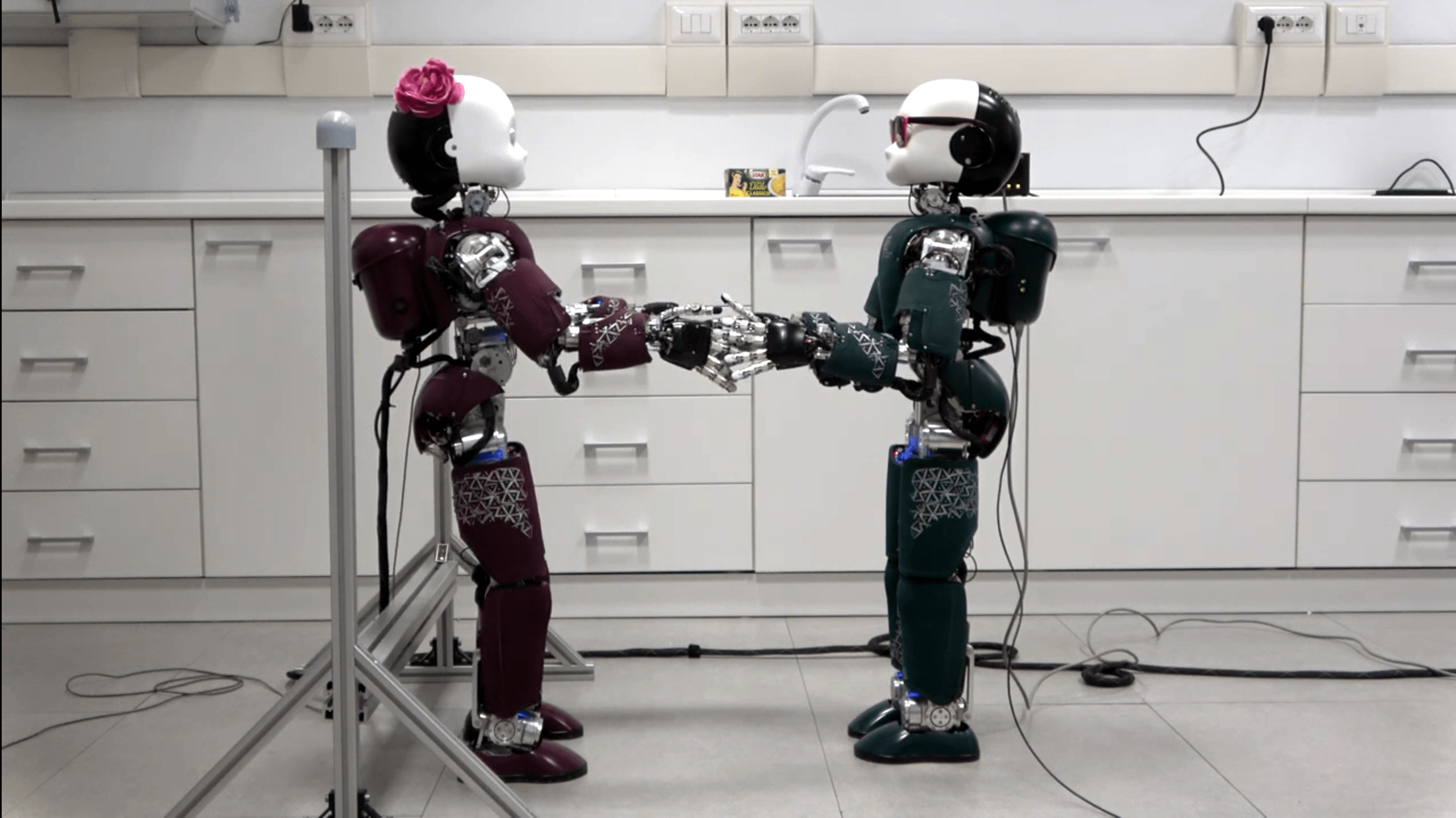}
		\caption{}
		\label{fig:state4}
	\end{subfigure}
	\caption{Stand-up experimental scenario with two iCub robots involved in physical interaction}
	\label{fig:two-icubs}
\end{figure*}

An experimental setup where two iCub humanoid robots are involved in a collaborative scenario with physical interactions is highlighted in Fig.\ref{fig:two-icubs}. The purple robot is torque controlled and is expected to perform the stand-up task with assistance from the green robot that is position controlled, executing pre-programmed movements that mimic pull-up assistance. The control laws are formulated by considering the state and the dynamics of both the robots at the modeling level. This facilitates the communication of the interacting agent's state and dynamics to the purple robot.

\begin{figure}[H]
    \centering
    \includegraphics[trim= 12cm 0cm 13cm 0cm, clip=true, scale=0.125]{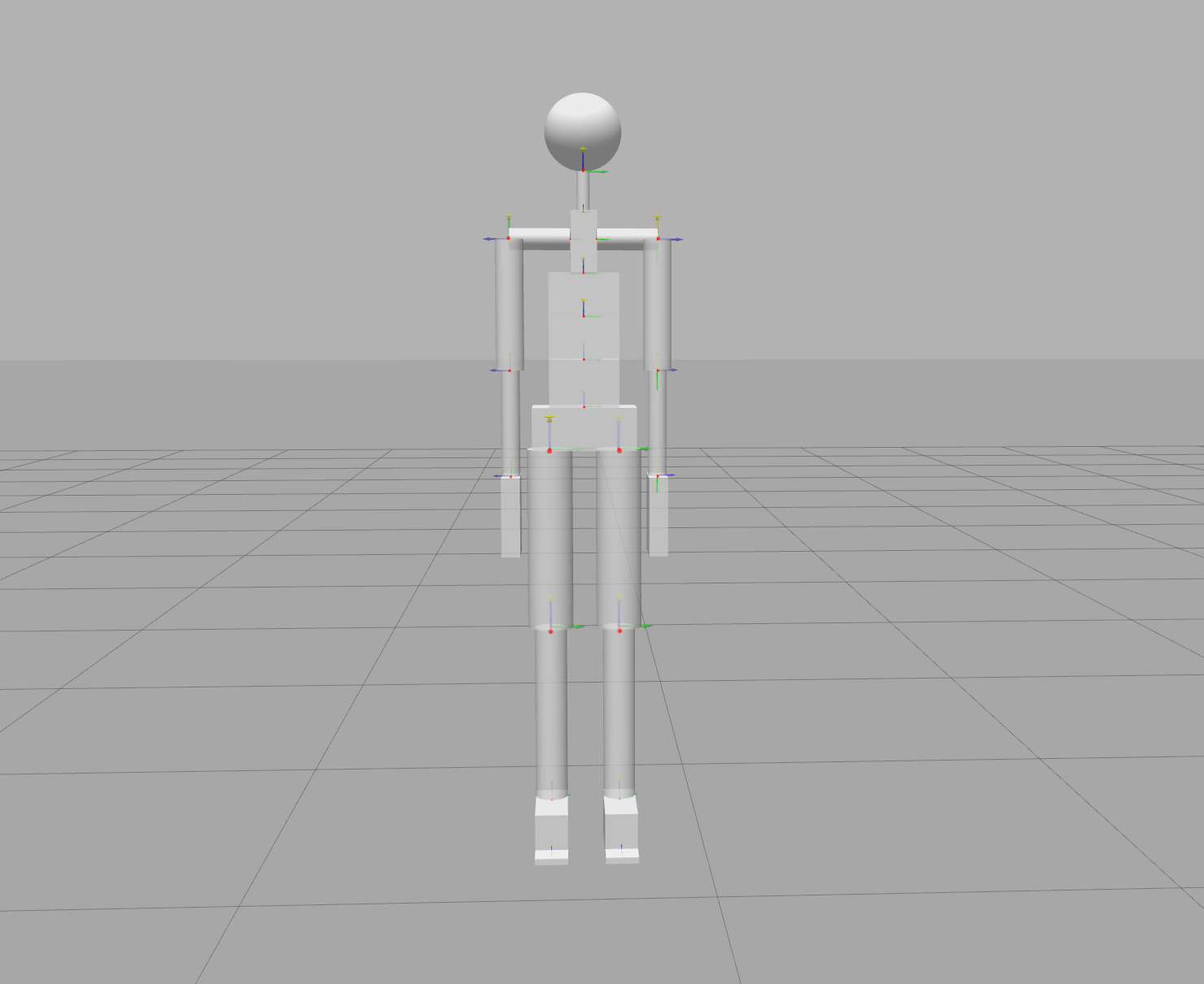}
    \caption{Human model as a multi-body mechanical system}
    \label{fig:human-model}
\end{figure}

At the modeling level, we can also consider humans as multi-body mechanical systems composed of rigid links that represent the properties of the human body segments. Fig.~\ref{fig:human-model} shows the model of a human as an articulated rigid body mechanical system. Although the assumption of a human body being modeled as rigid bodies is far from reality, it serves as a rough approximation to formulate HRC interaction dynamics and allows us to synthesize robot controllers optimizing both human and robot variables.

\begin{figure}[H]
    \centering
    \includegraphics[trim= 17.5cm 2.5cm 5cm 5cm, clip=true, scale=0.25]{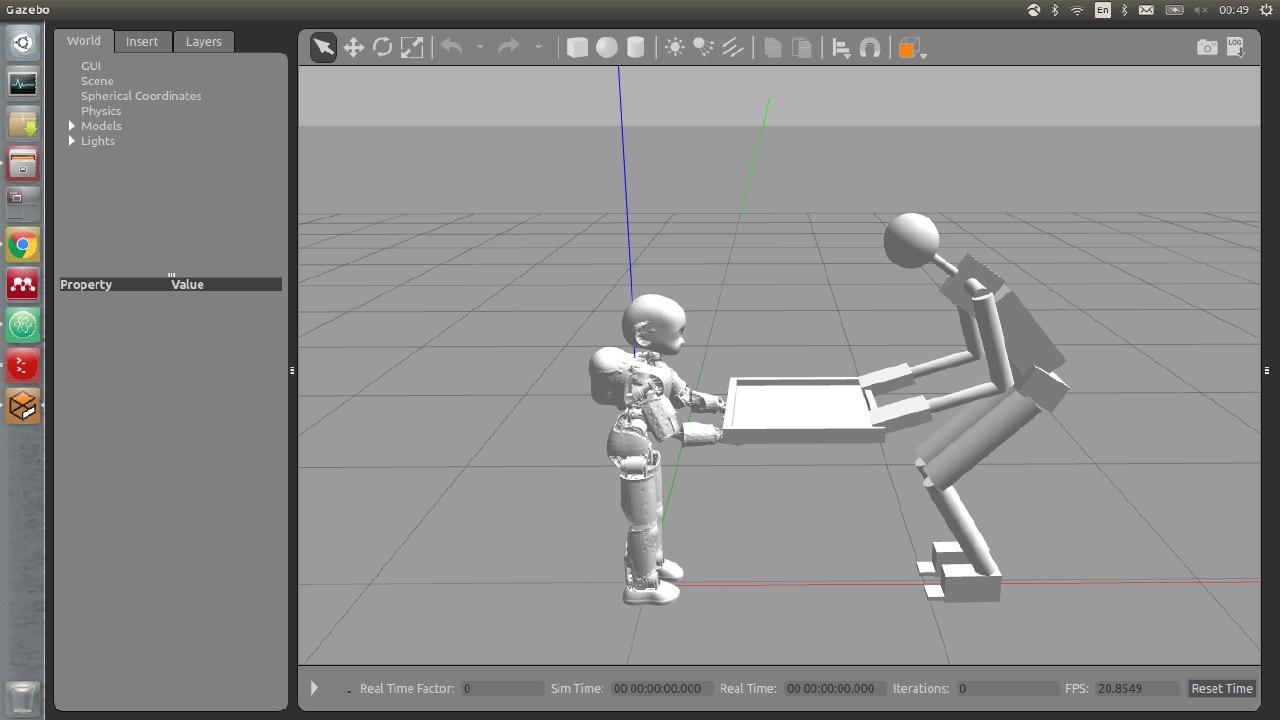}
    \caption{Human-Robot collaborative lifting task}
    \label{fig:human-robot}
\end{figure}

We are currently investigating partner-aware robot control techniques towards optimizing the ergonomy of the interacting agent during HRC scenarios. An example scenario of ergonomy optimization is a collaborative lifting task as shown in Fig.~\ref{fig:human-robot} where the humanoid robot needs to support the human.

\begin{figure}[H]
    \centering
    \includegraphics[scale=0.25]{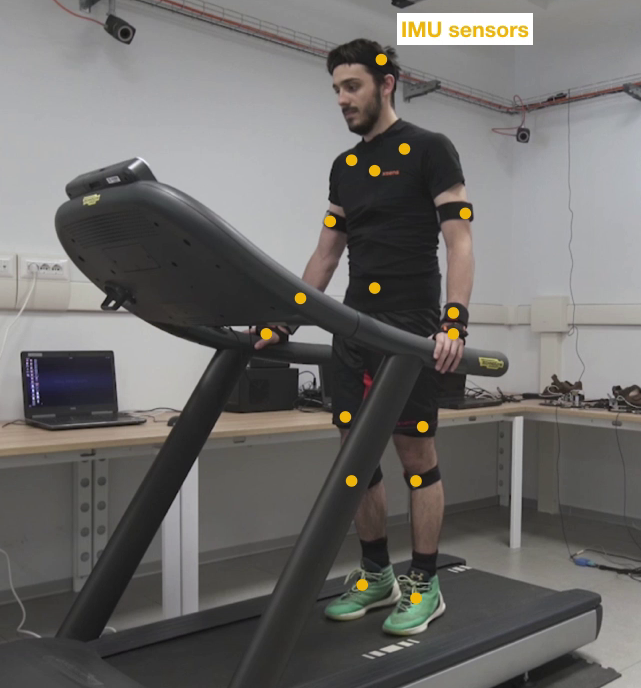}
    \caption{Human with distributed IMU sensors}
    \label{fig:human-xsens}
\end{figure}

\begin{figure}[H]
	\centering
	\begin{subfigure}{0.24\textwidth}
		\centering
        \includegraphics[scale=0.1775]{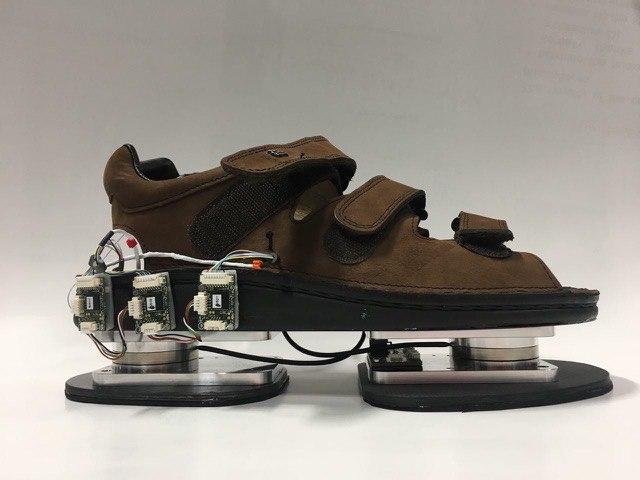}
        \caption{}
        \label{fig:ftshoes}
	\end{subfigure}%
	\begin{subfigure}{0.24\textwidth}
		\centering
		\includegraphics[scale=0.075]{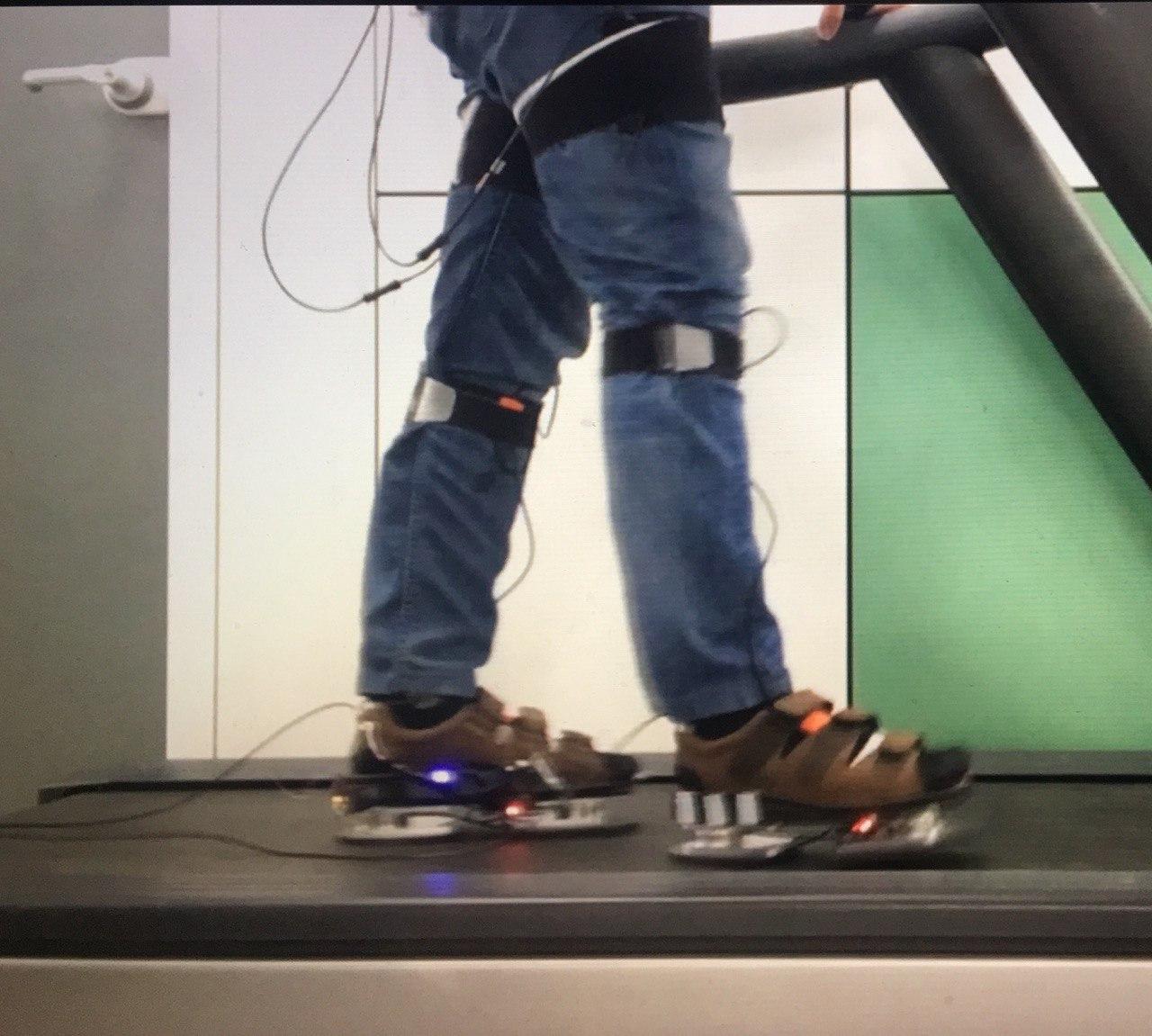}
		\caption{}
		\label{fig:human-ftshoes}
	\end{subfigure}
	\caption{Sensorized force/torque wearable shoes}
	\label{fig:human-dynamics}
\end{figure}

A key detail to realize such partner-aware control with humans is the capacity to acquire human data i.e. human joint positions, velocities, accelerations, and torques in real-time. Human motion tracking using computer vision techniques is a fairly advanced technology that can run even on a smart device. However, they are computationally intensive, often fail during occlusions and are not robust to provide real-time tracking for robotic applications. So, we focus on using a wearable suit of Inertial Motion Units (IMUs) distributed over the human body as shown in Fig.~\ref{fig:human-xsens} and developed real-time human motion tracking through dynamical inverse kinematics optimization for floating-base articulated systems as humans \cite{Rapetti2019}. Furthermore, we developed sensorized force/torque wearable shoes as shown in Fig.~\ref{fig:human-dynamics} that are leveraged to perform simultaneous floating-base estimation of human kinematics and joint torques \cite{latella2019simultaneous}.

\begin{figure}[H]
	\centering
	\includegraphics[scale=0.1]{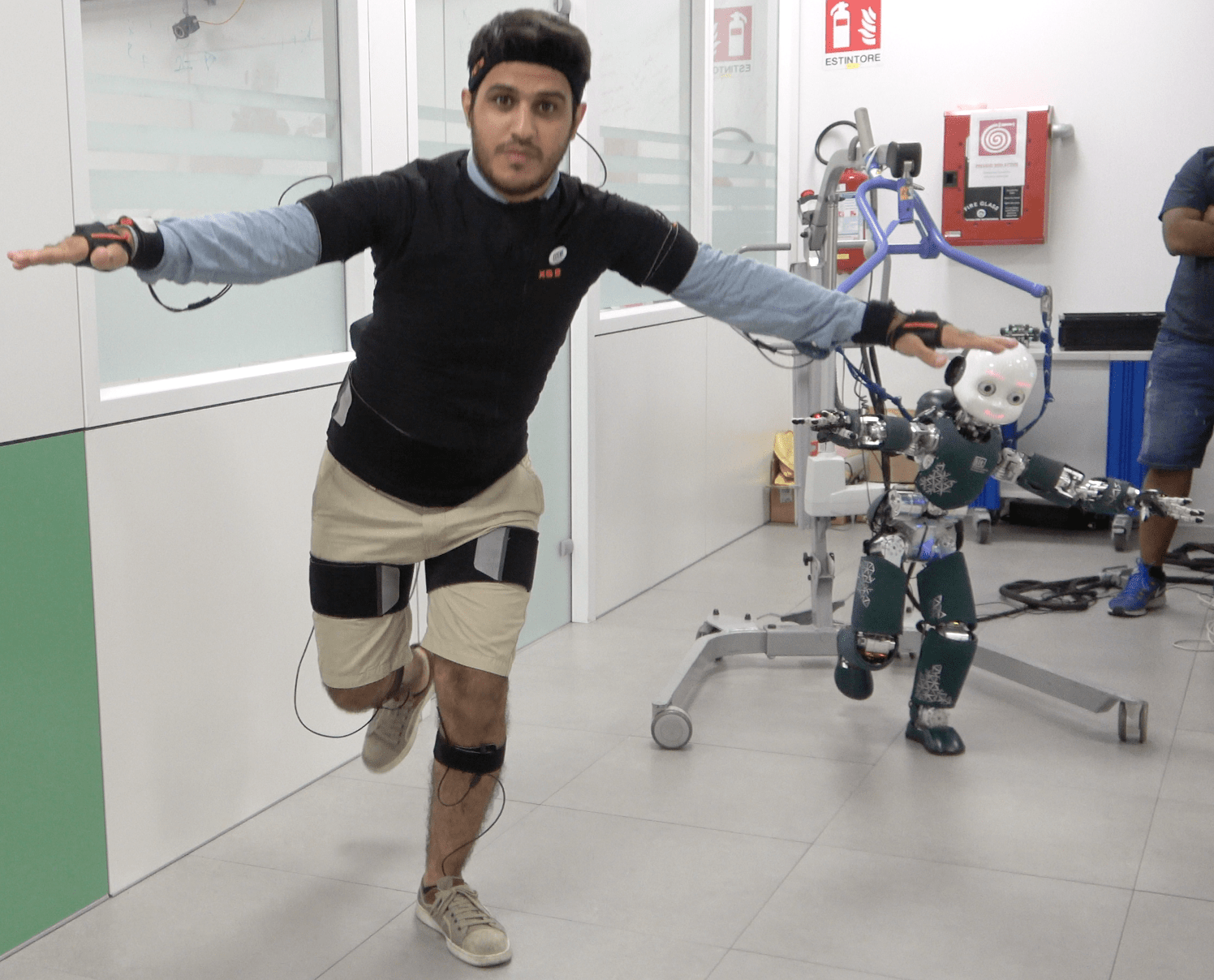}
	\caption{Whole-body retargeting example scenario}
	\label{fig:whole-body-retargeting-example}
\end{figure}

Another practical and interesting application to facilitate communication in human-robot teams towards joint action is whole-body human motion retargeting to humanoid robots and teleoperation \cite{whole-body-retargeting}. An example scenario of whole-body retargeting of human motion to a humanoid robot is shown in Fig.~\ref{fig:whole-body-retargeting-example} where each limb of the robot mimics the motion of the human limbs. Anthropomorphic motions from a human can be retargeted in real-time to a humanoid robot which will increase the trust in humans interacting with the robot.

\bibliographystyle{IEEEtran}
\bibliography{references}

\end{document}